\documentclass{article}

   \usepackage[preprint]{neurips_2020}

\usepackage[utf8]{inputenc} 
\usepackage[T1]{fontenc}    
\usepackage{hyperref}       
\usepackage{url}            
\usepackage{booktabs}       
\usepackage{amsfonts}       
\usepackage{nicefrac}       
\usepackage{microtype}      

\usepackage{color,soul}
\usepackage[]{listings}
\usepackage[]{mathcomp}
\usepackage{graphicx}
\title{A Modified Convolutional Network for Auto-encoding based on Pattern Theory Growth Function}

\author{%
  Erico Tjoa\thanks{Alibaba-NTU Joint Research Institute} \\
  Nanyang Technological University\\
  50 Nanyang Ave, 639798 \\
  \texttt{ericotjo001@e.ntu.edu.sg} \\

}

\begin{document}

\maketitle

\begin{abstract}
 This brief paper reports the shortcoming of a variant of convolutional neural network whose components are developed based on the pattern theory framework.
\end{abstract}

\section{Introduction}

The pattern theory has been extensively studied and well-documented (\cite{grenander1993general}). This document is a short report on an experimental extension to \cite{Tjoa2021ConvolutionalNN} that applies the abstract biological growth function introduced in the pattern theory on deep convolutional neural network (CNN). We loosely extend the concepts of generators and growth function into the realm of continuous parameters. Convolutional layers are arranged to into encoder/decoder blocks which are arranged to form a multi-layer neural network. While the ultimate objective is to study the interpretability of deep neural network by identifying meanings in variables encoded through the pattern theory framework, at this early stage, we focus first on the quality of the auto-encoding performance. The code is available\footnote{\url{https://github.com/etjoa003/gpt/tree/main/src/gpt_genmod}}.

\subsection{Related Works}
Variational Auto-Encoder (\cite{Kingma2014}) encodes each data point into a continuous vector of latent variables. It uses reparameterization trick to introduce normally distributed random values into the auto-encoder training pipeline. Many VAE variants have also been developed to improve its performance. This paper is a brief report regarding an experiment on an auto-encoding model whose encoder and decoder are based on the above-mentioned pattern theory concept. Reparameterization is simplified to uniform random noise injection to the latent distribution.

Generative adversarial network (GAN) (\cite{NIPS2014_5ca3e9b1}) has been well-known in the generative modelling community, where the generator model \(G_{GAN}\) generates data from normally distributed latent vector and the discriminative model \(D_{GAN}\) predicts if the data is real or artificial. \(G_{GAN}\) is trained to fool \(D_{GAN}\) into believing that data generated by \(G_{GAN}\) are real while \(D_{GAN}\) is trained to distinguish fake data from the real. Such adversarial training has produced highly realistic data, and similarly, many excellent variants have been developed. Fr\'{e}chet Inception Distance (FID), SSIM and MS SSIM have been used for evaluating the quality of dataset generated from a GAN. This report will use these values for evaluation as well.

Minimalist review on the general pattern theory can be found in \cite{Tjoa2021ConvolutionalNN}. The abstract biological growth process starts with (1) an initial configuration of generators, is followed by (2) the computation of an array of environment values ENV due to existing generators, and then (3) the evolution of the configuration according to a given rule. Repeat step 1 to 3 till the desired number of time steps. Types of generators and bond values of the generators used in chapter 4 of \cite{grenander1993general} and \cite{Tjoa2021ConvolutionalNN} are discrete integers. The experiment reported here modifies the process by creating a continuous generalization of generators and bond values.

\section{Method}
\begin{figure}[h!]
  \includegraphics[width=1.0\textwidth , trim = {0 0 0 0.1cm}]{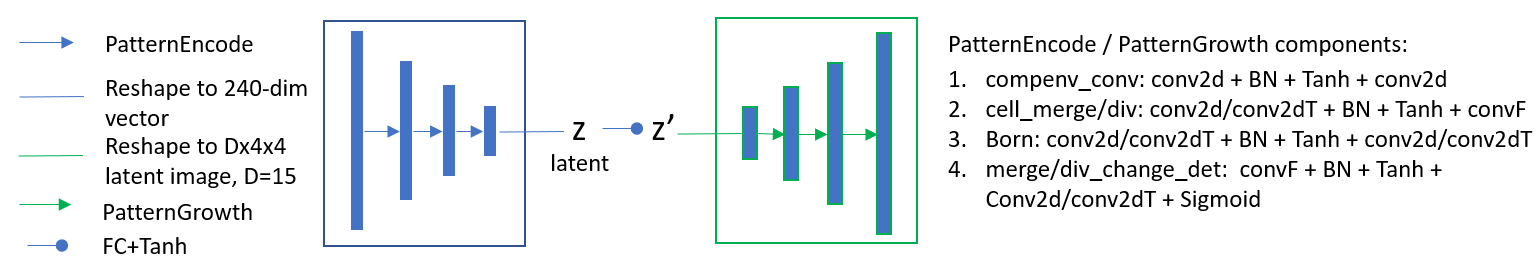}
  \caption{SimpleGrowth architecture. BN: batch normalization layer, conv2d: convolutional layer, conv2dT: transposed convolutional layer, ConvF: Conv2dT + Tanh + Conv2d + Tanh, used to add depth to CNN while maintaining feature map size without padding.}
  \label{fig:arch}
\end{figure}

CIFAR-10 dataset is trained on the SimpleGrowth architecture as shown in fig. \ref{fig:arch}. It is trained for 48 epochs with batch size 16, learning rate 0.001, using adam optimizer with \(\beta=(0.5,0.999)\) optimizing plain MSE loss. The encoder is arranged to encode each input image into a 240-dimensional latent vector. Encoder/decoder of the model consists of blocks of PatternEncode/PatternDecode that forward-propagates signals as shown in algorithm 1. SSIM and MS SSIM are measured on 4096 reconstructed CIFAR-10 images. Likewise, 4096 test images are compared to the same number of reconstructed images drawn from the test dataset for FID computation. We also train a similar architecture (with extra one encoder/decoder block) on CelebA 64x64 dataset for 2 epochs. 

At the fully connected (FC) layer where an image has been encoded into a vector with 240 dimensional latent variables, uniform random noise from \([-0.1,0.1]\) is injected as a simplified version of reparameterization trick used in VAE. The noise is added onto signals that have been passed through Tanh activation i.e. the original encoded signals are in the range of \([-1,1]\). The noise is intended to establish continuity in latent variables such that each point is robust to a radius of \(0.1\) in the latent space.

Algorithm 1 is an arbitrary choice of continuous generalization for GROWTH1 function found in chapter 4 of \cite{grenander1993general}. Rather than discrete bond values attached to each generator available in the generator space \(G\), convolutional layers are used. For example, compenv(x) in algorithm 1 shows how ENV is computed by the standard forward pass of a convolutional layer instead of drawing one of the generators from \(G\) with discrete values. 

\begin{minipage}{.55\textwidth}
\begin{lstlisting}[title={\textbf{Algorithm 1}},columns=fullflexible]
def forward(x):
    # PatternEnccode/Growth forward propagation.
    ENV = compenv(x)
    x = growth(x, ENV)
    return x    
    
def growth(x, ENV):
    type_and_env = concatenate((x[:,:DTYPE,:,:], ENV), dim=1)
    CHANGE = self.merge/div_change_det(type_and_env)
    x = cell_merge/div(x)
    born = born(x)
    x = x*(1-CHANGE) + CHANGE * born
    x[:,:3,:,:] = sigmoid(x[:,:3,:,:])
    return x

\end{lstlisting}
\end{minipage}\hfill
\begin{minipage}{.45\textwidth}
\begin{lstlisting}[columns=fullflexible]
def compenv(x):
    J = topology
    batch_size, c, h, w = x.shape
    ENV = zeros(batch_size, Nj, h, w)
    for k in range(Nj):
        roll_right = J[k,0]
        roll_down =  -J[k,1]
        temp = roll(x, int(roll_right), 1+2) 
        temp = roll(temp,int(roll_down),0+2) 
        ENV[:,k:k+1,:,:] = compenv_conv(temp)
    return ENV
 
\end{lstlisting}
\end{minipage}

\section{Results}
\begin{figure}[h!]
  \includegraphics[width=1.0\textwidth , trim = {0 0 0 0.1cm}]{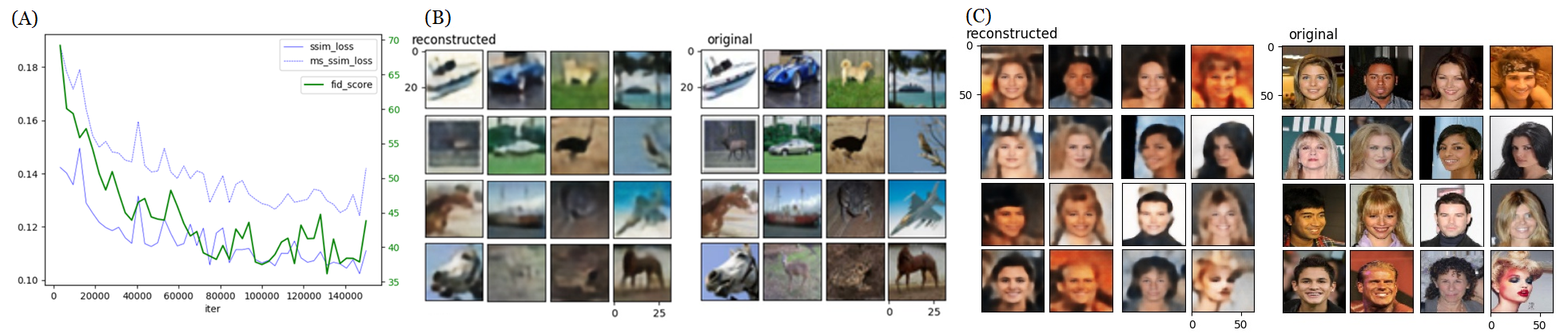}
  \caption{(A) MS SSIM, SSIM and FID measured on images reconstructed using SimpleGrowth architecture along the training iterations. (B) Reconstructed CIFAR-10 images compared to the original images drawn from the test dataset. (C) Reconstructed CelebA 64x64 images.}
  \label{fig:1}
\end{figure}

Unfortunately, the architecture does not yield considerable benefits apart from possibly fast qualitative convergence in the earlier training phase, when generated images can start to be visually recognized. The images after we complete the training are seen in fig. \ref{fig:1}(B,C), where CelebA images are reconstructed after only 2 epochs. The best FID achieved for CIFAR-10 attained during 48 epochs was 36.15 although FID as low as 2.2 has been achieved in \cite{Song2020ScoreBasedGM}. At the end of epoch 2 for CelebA 64x64 training, FID value of 86.2 is obtained though FID of 5.25 has been achieved in \cite{Aneja2020NCPVAEVA}. Note: we compute FID between all test dataset and training dataset available in CIFAR-10, obtaining a value of 3.11.

\begin{figure}[h!]
  \includegraphics[width=1.0\textwidth , trim = {0 0 0 0.1cm}]{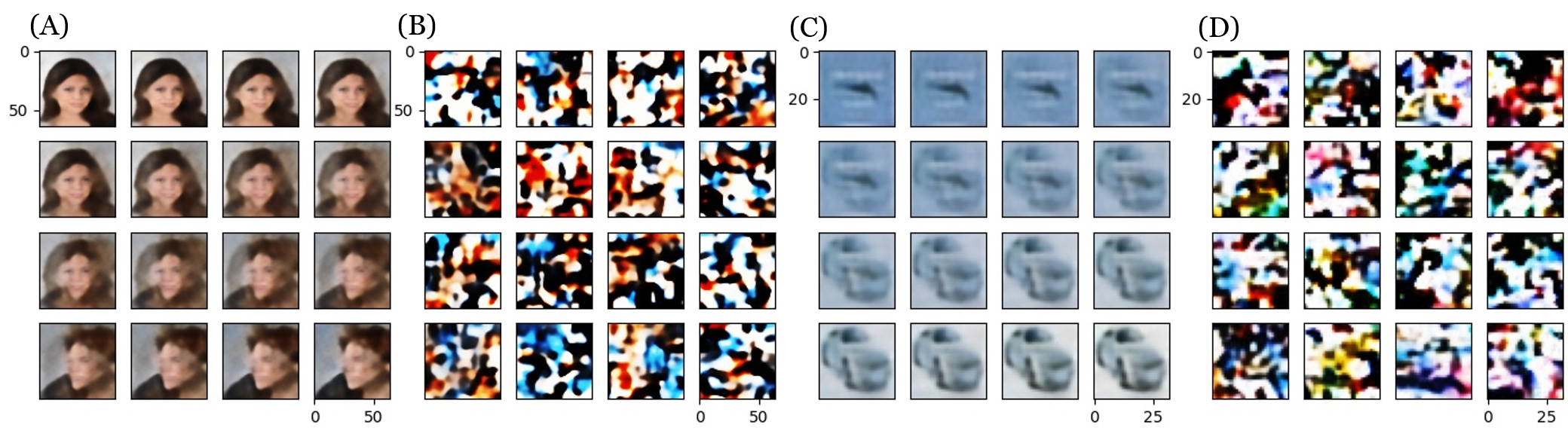}
  \caption{(A) Two images reconstructed from CelebA 64x64 training dataset with latent variable transition. (B) Images decoded latent vector drawn from uniform random distribution \([-1,1]\). (C,D) are similar to (A,B), but for CIFAR-10.}
  \label{fig:2}
\end{figure}

Fig. \ref{fig:2}(A) is obtained from the linear transition of two reconstructed images \(x_1,x_2\) (top-left and bottom-right of the image). Their latent variables \(z_1,z_2\) are computed through the encoder layer of SimpleGrowth, and then all the 16 images are obtained by decoding the latent vector \(z_k=z_1+ k\times (z_2-z_1)/16\) for \(k=0,1,\cdots,15\), i.e. latent variables are transitioned linearly along all dimensions. The transition does not reflect any meaningful transition. Fig. \ref{fig:2}(C) shows the transition from an airplane to a car. Likewise, direct juxtaposition between images are observed in the transition stage, showing no meaningful transformation of visual components. The architecture does not achieve the ``semantic continuity" of latent variables. Furthermore, latent vectors drawn from \([-1,1]\) uniform random distribution are decoded into visually unrecognizable images as shown in fig. \ref{fig:2}(B,D) for CIFAR-10 and CelebA 64x64 respectively.

\textit{Remark}. Naturally, when we increase the number of latent dimensions, the quality of images improved (not reported). This is counter-productive since the encoding will be more spatially inefficient.

\section{Conclusion}
This paper generally reports the poor quality of encoding/decoding performance of a convolutional neural network designed with propagation method based on the growth function defined in the pattern theory. As far as we know, there is not yet a canonical way to generalize the growth function and systems of generator configurations. There are still many different possibilities for development along this relatively unexplored direction. The main question remains, is there a theoretically sound architecture based on pattern theory framework that will greatly improve the auto-encoding quality? Other future directions that can be studied piecewise include (1) reparameterization: what is the radius of uniform random variables required to create a meaningful continuous latent distribution (2) can we improve the quality of SimpleGrowth architecture using improvements that have been developed for the VAE?

\section*{Broader Impact}
This is an early stage of an effort to study of interpretability of the deep neural network through pattern theory framework. Interpretability of deep neural network enables more transparent and fairer use of algorithms.

\begin{ack}
This study is conducted under NTU-Alibaba Talent Program.
\end{ack}

\bibliographystyle{unsrtnat}
\bibliography{genmodarx}

\begin{thebibliography}{6}
\providecommand{\natexlab}[1]{#1}
\providecommand{\url}[1]{\texttt{#1}}
\expandafter\ifx\csname urlstyle\endcsname\relax
  \providecommand{\doi}[1]{doi: #1}\else
  \providecommand{\doi}{doi: \begingroup \urlstyle{rm}\Url}\fi

\bibitem[Grenander(1993)]{grenander1993general}
Ulf Grenander.
\newblock \emph{General Pattern Theory: A Mathematical Study of Regular
  Structures}.
\newblock Oxford Mathematical Monographs. Clarendon Press, 1993.
\newblock ISBN 9780198536710.
\newblock URL \url{https://books.google.com.sg/books?id=Z-8YAQAAIAAJ}.

\bibitem[Tjoa and Guan(2021)]{Tjoa2021ConvolutionalNN}
Erico Tjoa and Cuntai Guan.
\newblock Convolutional neural network interpretability with general pattern
  theory.
\newblock \emph{ArXiv}, abs/2102.04247, 2021.

\bibitem[Kingma and Welling(2014)]{Kingma2014}
Diederik~P. Kingma and Max Welling.
\newblock {Auto-Encoding Variational Bayes}.
\newblock In \emph{2nd International Conference on Learning Representations,
  {ICLR} 2014, Banff, AB, Canada, April 14-16, 2014, Conference Track
  Proceedings}, 2014.

\bibitem[Goodfellow et~al.(2014)Goodfellow, Pouget-Abadie, Mirza, Xu,
  Warde-Farley, Ozair, Courville, and Bengio]{NIPS2014_5ca3e9b1}
Ian Goodfellow, Jean Pouget-Abadie, Mehdi Mirza, Bing Xu, David Warde-Farley,
  Sherjil Ozair, Aaron Courville, and Yoshua Bengio.
\newblock Generative adversarial nets.
\newblock In Z.~Ghahramani, M.~Welling, C.~Cortes, N.~Lawrence, and K.~Q.
  Weinberger, editors, \emph{Advances in Neural Information Processing
  Systems}, volume~27. Curran Associates, Inc., 2014.
\newblock URL
  \url{https://proceedings.neurips.cc/paper/2014/file/5ca3e9b122f61f8f06494c97b1afccf3-Paper.pdf}.

\bibitem[Song et~al.(2020)Song, Sohl-Dickstein, Kingma, Kumar, Ermon, and
  Poole]{Song2020ScoreBasedGM}
Yang Song, Jascha Sohl-Dickstein, Diederik~P. Kingma, A.~Kumar, S.~Ermon, and
  Ben Poole.
\newblock Score-based generative modeling through stochastic differential
  equations.
\newblock \emph{ArXiv}, abs/2011.13456, 2020.

\bibitem[Aneja et~al.(2020)Aneja, Schwing, Kautz, and
  Vahdat]{Aneja2020NCPVAEVA}
Jyoti Aneja, Alexander~G. Schwing, J.~Kautz, and Arash Vahdat.
\newblock Ncp-vae: Variational autoencoders with noise contrastive priors.
\newblock \emph{ArXiv}, abs/2010.02917, 2020.

\end{thebibliography}

%
%
%
%

\end{document}